\documentclass{article}
\usepackage{hyperref}
\usepackage{cleveref}
\usepackage[T1]{fontenc}
\usepackage[utf8]{inputenc}
\usepackage{authblk}

\title{An unsupervised learning approach to solving heat equations on chip based on Auto Encoder and Image Gradient}

\author[a, \thanks{Corresponding author, email: haiyang.he@ansys.com (Haiyang He)}]{Haiyang He}
\author[b]{Jay Pathak}
\affil[a]{Ansys Inc., Canonsburg, Pennsylvania, USA}
\affil[b]{Ansys Inc., San Jose, California, USA}

\date{July 2020}

\usepackage[square,numbers]{natbib}
\usepackage{graphicx}

\begin{document}

\maketitle

\begin{abstract}
    Solving heat transfer equations on chip becomes very critical in the upcoming 5G and AI chip-package-systems. However, batches of simulations have to be performed for data driven supervised machine learning models. Data driven methods are data hungry, to address this, Physics Informed Neural Networks (PINN) have been proposed. However, vanilla PINN models solve one fixed heat equation at a time, so the models have to be retrained for heat equations with different source terms. Additionally, issues related to multi-objective optimization have to be resolved while using PINN to minimize the PDE residual, satisfy boundary conditions and fit the observed data etc. Therefore, this paper investigates an unsupervised learning approach for solving heat transfer equations on chip without using solution data and generalizing the trained network for predicting solutions for heat equations with unseen source terms. Specifically, a hybrid framework of Auto Encoder (AE) and Image Gradient (IG) based network is designed. The AE is used to encode different source terms of the heat equations. The IG based network implements a second order central difference algorithm for structured grids and minimizes the PDE residual. The effectiveness of the designed network is evaluated by solving heat equations for various use cases. It is proved that with limited number of source terms to train the AE network, the framework can not only solve the given heat transfer problems with a single training process, but also make reasonable predictions for unseen cases (heat equations with new source terms) without retraining.
\end{abstract}

\section{Introduction}
Machine learning has been growing exponentially and achieving tremendous success in various applications in recent few years\citep{he2016joint}, such as image classification, pattern recognition, question answering\citep{xia2018zero}, robotics, advanced manufacturing systems\citep{he2019machine, huang2020unsupervised, yang2019new} and so on. Artificial neural networks and other machine learning algorithms can be extensively used for predicting weather, calamities, severe illnesses and facilitating research on medical, autonomous vehicles and recommendation systems. 

Although the concept of using neural networks to solve partial differential equations (PDEs) goes back to 1990s \citep{psichogios1992hybrid, dissanayake1994neural, lagaris1998artificial}, more extensive and active research on it emerges in recent few years. Based on whether training data is required, these approaches can be classified into supervised and unsupervised genres. For supervised approach, a data-driven framework for learning unknown time-dependent autonomous PDEs using deep neural networks is presented in \citep{wu2020data}. On the other hand, the function approximation capabilities of feed-forward neural networks are utilized to provide accurate solutions to ordinary and partial differential equations in an unsupervised way in \citep{shirvany2009multilayer}. In terms of network architectures, feed forward neural networks is widely used. For example, a method for constructing multilayer neural network approximations of solutions of differential equations, based on the finite difference method, is proposed in \citep{kaverzneva2019differential}. A multilayer neural networks is used in \citep{he2000multilayer} and an extended backpropagation algorithm was proposed to approximate solution for a class of first-order partial differential equations for input-to-state linearizable or approximate linearizable systems. A “Deep Galerkin Method (DGM)” based deep neural network is trained to solve high-dimensional PDEs in \citep{sirignano2018dgm} and is able to solve PDEs up to 200 dimensions. Besides, there has been researches using Convolutional Neural Network (CNN) based structure as well, for example, a U-Net architecture is used in \citep{winovich2019convpde} and a CNN-based encoder-decoder model is used in \citep{ranade2020discretizationnet} for constructing numerical solvers for PDEs. With recent developments in one of the most useful but perhaps underused techniques in scientific computing: automatic differentiation (AD) \citep{baydin2017automatic}, the idea of defining partial differential equations utilizing AD, imposing physics through neural network losses together with easy access to backpropagation revolutionized PDE solving by using Physics-Informed Neural Networks (PINN). Implementations of PINN on solving PDEs are investigated in \citep{raissi2017physics, raissi2018forward, raissi2019physics, lu2019deepxde}.

Naturally, neural networks has been investigated in solving parabolic PDEs as well, such as heat equations. For instance, a continuous-time analogue Hopfield neural network (CHNN) is proposed in \citep{deng2006applying} to solve time-varying forward heat conduction problems. PDE model problems including heat equations are solved using data driven based artificial neural networks in \citep{mishra2018machine}. Advection and diffusion type PDEs in complex geometries are solved by modifying backpropagation algorithm in \citep{berg2018unified}. Heat equations are solved by deep neural networks and Automatic Differentiation based framework in \citep{liu2019neural}. With the rapid development of solving heat equations using neural networks, given one specific problem, most of the approaches mentioned above would obtain desirable predictions of the corresponding solutions. However, such a trained network would struggle if the inference problem is slightly modified, e.g. a new source term is introduced to the heat equation. Usually the network needs to be retrained for the unsupervised models or more solution data should be collected for retraining the supervised algorithms, which diminishes the efficiency and effectiveness of the trained neural network.

In addition, it is of critical importance to investigate improving the generalization capability of neural network to different source terms for the heat equations. For instance, chip thermal analysis (CTA) becomes important for the performance and reliability of chip-package-system in the upcoming 5G and AI systems \citep{Jimin, JiminGTC}. One interesting problem is obtaining the temperature profile on a chip given a certain power map, which can be treated as the source term in heat equations. Conventionally, finite
element method (FEM) or computational fluid dynamics
(CFD) technology is applied to acquire an accurate thermal profile on-chip and is very time-consuming. Not to mention that in CTA problems the source term will change when the design of the chip or input to the chip alters so the heat transfer problem to be solved will vary accordingly. Thus, large amount of numerical simulations have to be performed, which is labour-intensive. To alleviate the problem, a data driven DNN-based approach is proposed in \citep{Jimin, JiminGTC} and it can achieve run time reduction of greater than 100-1000x with good accuracy compared to the traditional FEA based method. However, data-driven approaches usually require generating the solutions to the problems in advance for training the models, which will still incur batches of simulations. Unsupervised models such as PINN \citep{raissi2017physics} does not require a lot of training data whereas a vanilla PINN trained for one heat source is not expected to make satisfactory inference on another heat source so it has to be retrained for every unseen source term. Currently, there is very few research papers investigating generalizing a trained neural network based heat equation solver to various source terms (seen and unseen) at the same time, especially for unsupervised approaches. 

Against this background, a hybrid framework which integrates an Auto Encoder (AE) and Image Gradient (IG) based network is proposed. The task of the AE is to learn and compress the PDE source terms to latent vectors, which will be input to the IG based network. The PDE residual, which is designed as the loss function of the IG based network, will be minimized through network training. The hybrid framework on the one hand can focus on minimizing the PDE residuals for various heat equations at the same time, on the other hand, it will associate the residual minimization with these compact representations of source terms (AE latent vectors). It is shown through test cases that this framework can handle solving the heat equations and learning the generalization at the same time. 

This paper is organized as follows. In Section 2, the CTA problem is defined and the theoretical foundation of the proposed framework is explained. The hybrid framework of AE and IG based network is presented in Section 3.  Use cases for demonstrating the capability of the framework on solving heat equations and making predictions for unseen power maps are provided in Section 4.  Finally, conclusions are given in Section 5.

\section {Solving CTA using neural network} 
This section first introduces the CTA problem this paper aims at solving and then discusses neural network as a function approximator. The presented neural network framework is approximating a function mapping input power maps to heat equation solutions (temperature distributions).

\subsection{CTA}

\begin{figure}[htbp!]
\centering
\includegraphics[scale=0.4]{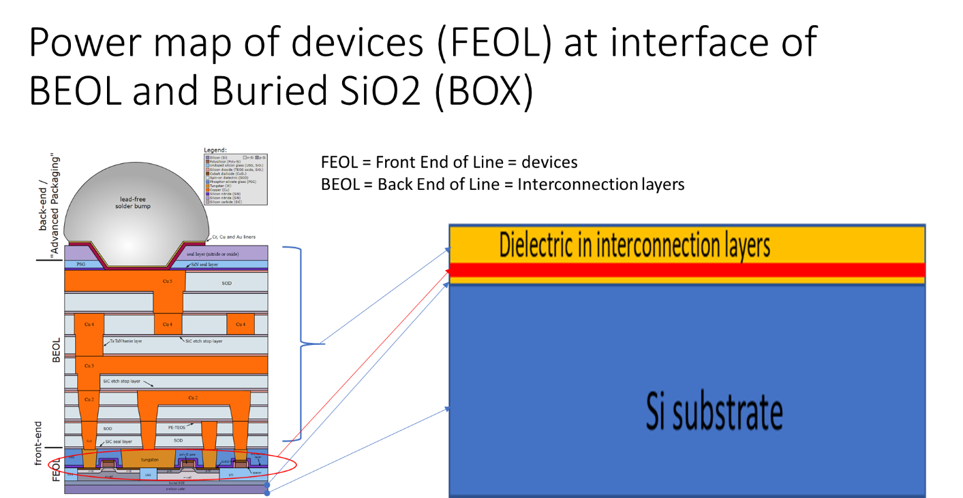}
\caption{Side view of a chip package. Left: side view of the whole package, right: details of the die model}
\label{fig:Chip_package}
\end{figure}

A side view of  a chip is given in Fig. \ref{fig:Chip_package}, where the red solid line in the detailed die model (the right sub-figure) represents the power map side view.

\begin{figure}[htbp!]
\centering
\includegraphics[scale=0.4]{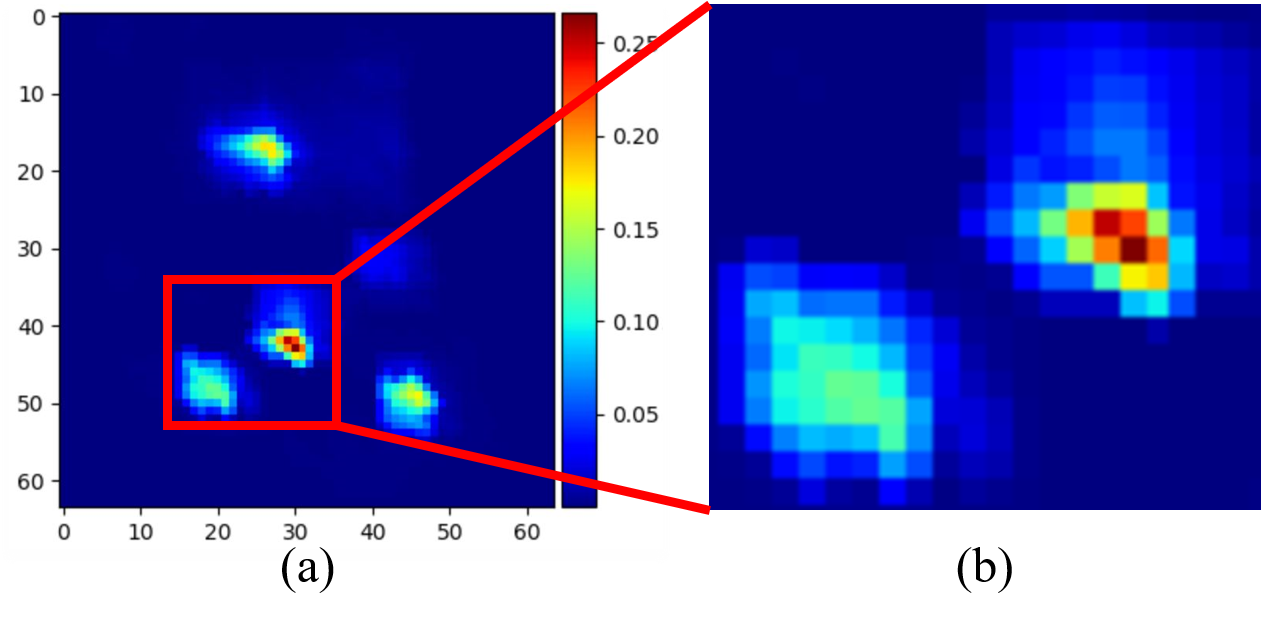}
\caption{Tile based power map: (a) power map example (b) Zoom in view of power region}
\label{fig:Example power map}
\end{figure}

In system-level thermal analysis using finite element method (FEM) or computational fluid dynamics (CFD) method, chip(s) are usually modeled as pure silicon blocks. The silicon block is further segmented into uniformly distributed tiles and the power map is considered as the general thermal environment of a chip. A tile-based power map can  usually be obtained from a dynamic voltage drop analysis tool. An example of tile based power map is shown in Fig. \ref{fig:Example power map}. In this paper, the focus will be on the temperature prediction of the chip given the device power map. The silicon block is simplified to be a 2D square or a 3D cube and the power map is applied on top of the square or in the center plane of the cube.

Artificially created power maps together with one realistic power map are investigated in this paper. The realistic power map is extracted from ANSYS RedHawk, which can import chip package layout and perform accurate on-chip static IR drop and AC hot spot analyses. 

\subsubsection{CTA in 2D}
In this study, the steady state temperature profile on each tile of the chip is of interesting. And the steady state  two-dimensional diffusion equation is given by Eq. (\ref {steady state diffusion}).

\begin{equation}
\frac{\partial^2 T}{\partial x^2} + \frac{\partial^2 T}{\partial y^2} = 0
\label{steady state diffusion}
\end{equation}

The power map applied on the chip (external heat source) is modeled as the source term of the heat equation, which is denoted as Q(x, y). Therefore, the 2D heat equation to be solved by the neural network is:

\begin{equation}
k(\frac{\partial^2 T}{\partial x^2} + \frac{\partial^2 T}{\partial y^2}) + Q(x, y) = 0
\label{heat equation with source}
\end{equation}
where $k$ is the thermal conductivity.

\subsubsection{CTA in 3D}
\begin{figure}[htbp!]
\centering
\includegraphics[scale=0.35]{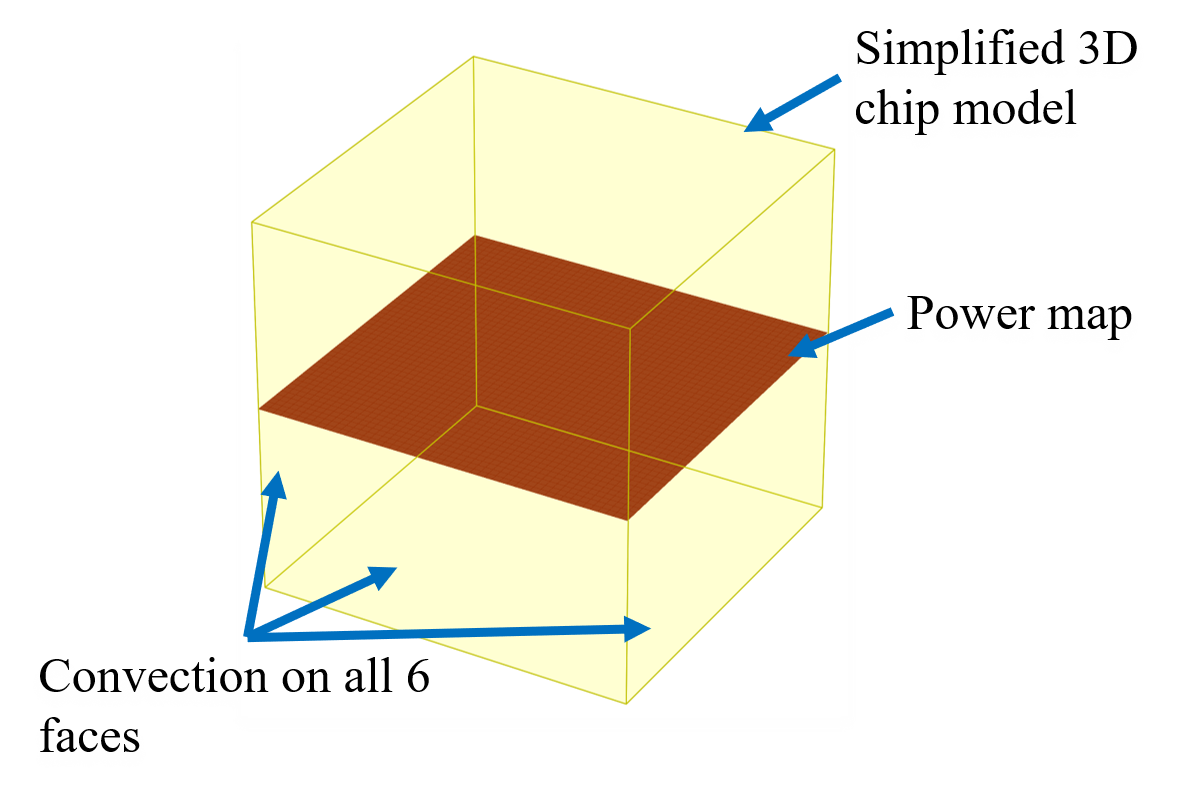}
\caption{Simplified 3D CTA}
\label{fig: Simplified 3D CTA}
\end{figure}

For conducting CTA in 3D, the chip is simplified and modelled as a 3D cube, a 2D power map is placed at its center plane and there is convection on the 6 faces of the cube, as shown in Fig. \ref{fig: Simplified 3D CTA}.

Therefore, the corresponding steady state 3D heat equation with convection is:

\begin{equation}
k(\frac{\partial^2 T}{\partial x^2} + \frac{\partial^2 T}{\partial y^2}+ \frac{\partial^2 T}{\partial z^2}) -(V_x\frac{\partial T}{\partial x}+V_y\frac{\partial T}{\partial y}+V_z\frac{\partial T}{\partial z})+ Q(x, y, z) = 0
\label{heat equation with source 3D}
\end{equation}
where $k$ is the thermal conductivity, $V_x$, $V_y$ and $V_z$ are constant velocities along $x$, $y$ and $z$ directions respectively and $Q(x, y, z)$ is the source term. 

\subsection{Neural network approximation theorem}
Based on the mathematical principle of neural networks, the universal approximation theorem \citep{csaji2001approximation} states that any continuous functions can be approximated by neural networks with non polynomial activation functions if no constraint is imposed on the number of hidden layers and the number of neurons in each layer \citep{hornik1989multilayer}. 

For a width-unbounded neural network with a single hidden layer \citep{haykin1994neural, hassoun1995fundamentals}, we have:
\begin{equation}
F(x) = \sum_{i=1}^{N}c_i\sigma(w_i^Tx + b_i)
\label{universal approximation theorem}
\end{equation}
where $\sigma$ is the activation function. Let $I_n$ represent closed n-dimensional unit hyper cube $[0, 1]^n$. And all continuous functions defined on $I_n$ are denoted as $C(I_n)$. Any continuous function defined in the domain of $C(I_n)$, which is $f \in C(I_n)$ can be accurately approximated with a fully connected neural network. The approximator  is guaranteed to have a small approximation error. That is, for any  $\varepsilon$ >0 and $x \in I_n$
\begin{equation}
\left| F(x) - f(x) \right|< \varepsilon
\end{equation}
Based on this theory, deep neural networks can be designed to approximate Eq. (\ref{heat equation with source}) and Eq. (\ref{heat equation with source 3D}) in this study.

Conventional neural network function approximator is constructed in a supervised learning process. The training data consists of variables $x_i$ and function values $y_i$. For a function approximator $\hat{y}$ at $x_i$, the loss function is defined as:
\begin{equation}
L = \sum_{i=1}^{N}\left| \hat{y}(x_i) - y_i \right|^2
\label{supervised loss}
\end{equation}
However, since an unsupervised approach is implemented in this study, the loss function is defined as the residual of function $f$ in Eq. \ref{res loss}. The optimization goal is to minimize the mean squared error of heat equation residual.
\begin{equation}
MSE_f = \frac{1}{N_f}\sum_{i=1}^{N_f}\left| f(x_f^i) \right|^2
\label{res loss}
\end{equation}
The details of the proposed framework are provided in Section \ref{framework}.

\section {The hybrid framework of AE and IG based network} \label{framework}
Principal component analysis (PCA) is one of the
most powerful and popular dimension reduction techniques, which can represent high-dimensional data with low-dimensional space. An autoencoder is a feed-forward neural network which is trained to map from a vector of values to the same vector \citep{kramer1991nonlinear}, which acts as a nonlinear implementation of a PCA using neural network architecture.  Due to the benefit of being efficient, flexible and versatile, autoencoders are widely used in compressing the input data.  It has shown some successful applications in the field of parameter estimation of PDEs \citep{rudy2019data, lu2019extracting, long2017pde, erichson2019physics,goh2019solving, pakravan2020solving, champion2019data}. Therefore, a convolutional AE is designed to compress the input power maps of the heat equations to be compact latent vectors.

An image gradient represents a directional variation in the intensity or color of an image. The gradient of a function of two variables at each image point is a 2D vector consisting of the derivatives in the horizontal and vertical directions. Thus, obtaining the derivative approximations of an image T with respect to spatial variables x and y, which are $\frac{\partial T}{\partial x}$ and $\frac{\partial T}{\partial y}$, using image gradient becomes straightforward. The working principle behind image gradient inspires the proposed IG based network. Intuitively, the temperature distribution profile in Eq. (\ref{heat equation with source}) or Eq. (\ref{heat equation with source 3D}) can be treated as an image, which is a function of spatial variables x and y. Therefore, to get the second order derivative approximations for  $\frac{\partial^2 T}{\partial x^2}$ and $\frac{\partial^2 T}{\partial y^2}$ in heat equations, we can make use of the output from image gradient function and realize an implementation of second order central differences.

Further information about the AE and IG based network are illustrated in Section \ref{AE} and Section \ref{IG} respectively. The integration of these two networks will be discussed in Section \ref{AEIG}.

\subsection{Auto encoder and Image Gradient based network}
\subsubsection{AE structure\label{AE}}
\begin{figure}[htbp!]
\centering
\includegraphics[scale=0.4]{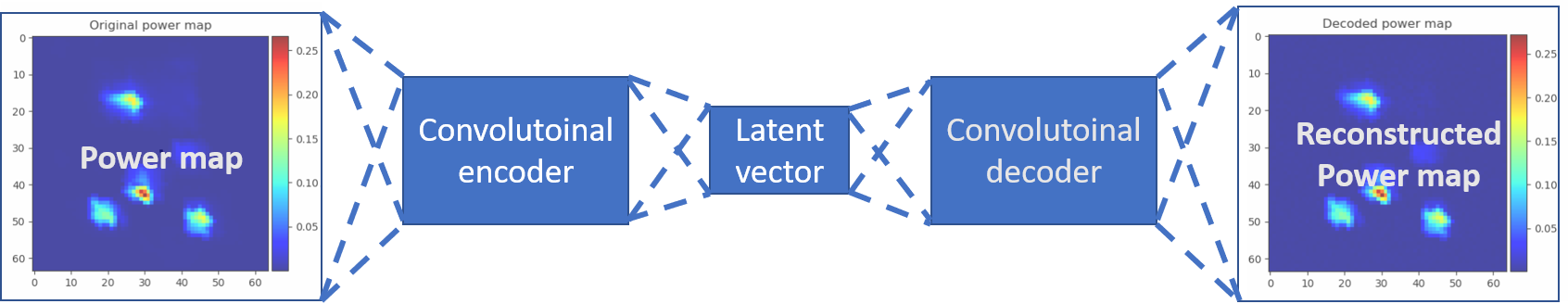}
\caption{Schematic of AE}
\label{fig: Schematic of AE}
\end{figure}

Since our inputs are power maps on a chip, which can be treated as images, it is straightforward to use convolutional neural networks as encoders and decoders. The implemented encoder consists of stacked  Conv2D and MaxPooling2D layers. The max pooling is used for spatial down-sampling on the power map. Correspondingly, the decoder consists of stacked  Conv2D and UpSampling2D layers. The power map is compressed to be a Dense layer. Power maps of shape $(batch size, height, width, channel)$ are input to the AE for training. The schematic of the AE is illustrated in Fig. \ref{fig: Schematic of AE}.

\subsubsection{Data generation and training of AE using artificial power maps}\label{section: artificial_power_map}

Different distributions of artificial power maps are created as the training data sets for AE, including sinusoidal power maps, exponential power maps, variations of one realistic power maps for 2D and 3D. For sinusoidal and exponential power maps, the input to AE can be treated as samples from continuous functions in 2D controlled by function parameters. For variations of one realistic power maps for 2D and 3D, artificial power maps are generated by image augmentation techniques (rotations and shifts) applied on one realistic power map from a chip design, as the one shown in Fig. \ref{fig: Schematic of AE} and Fig. \ref{fig: IG based network}. Mean Squared Error (MSE) of the residual is defined as the loss, Adam optimizer with learning rate of 1e-4 is used for optimization and the model is trained for 5000 epochs. 

\subsubsection{Image Gradient based network for solving the PDEs}\label{IG}

\begin{figure}[htbp!]
\centering
\includegraphics[scale=0.35]{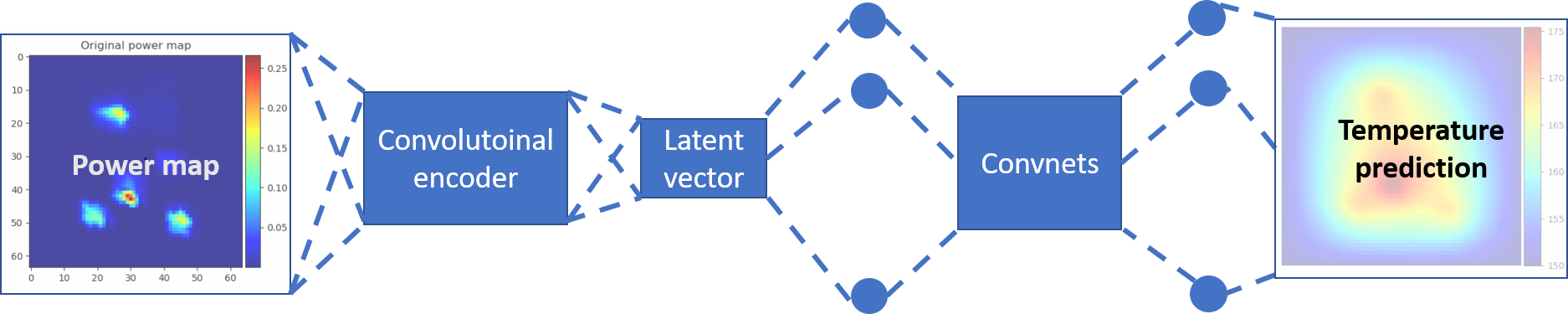}
\caption{IG based network}
\label{fig: IG based network}
\end{figure}

After the training of AE is finished, the encoder model will be extracted for encoding the power maps into compact vectors. These latent vectors will be fed to the IG based network for solving their corresponding PDEs. Convolutional layers are used in the IG based network, which is illustrated in Fig. \ref{fig: IG based network}.

\subsubsection{Training of IG}
The training of the network does not need solution data for the PDEs, instead, it will solve the heat equations by minimizing the PDE residual based on the second derivative central difference defined in Eq. \ref{IG_equation}, which is implemented by modifications of conventional image gradient function.

\begin{equation}
f''(x) \approx \frac{f(x + 2h) - 2f(x) + f(x - 2h)}{4h^2} 
\label{IG_equation}
\end{equation}

The boundary condition for all the power maps discussed in this paper is 150 $C$. It is encoded by padding the temperature prediction profile from the neural network with the boundary temperature values of the chip before performing the gradient calculation at each iteration.

The activation function used in the convolutional layers is tanh. Adam optimizer of learning rate of 1e-4 is used for network optimization. The model is trained for 10000 epochs to minimize the MSE loss of PDEs' residual.

\subsubsection{AEIG framework\label{AEIG}}
\begin{figure}[htbp!]
\centering
\includegraphics[scale=0.5]{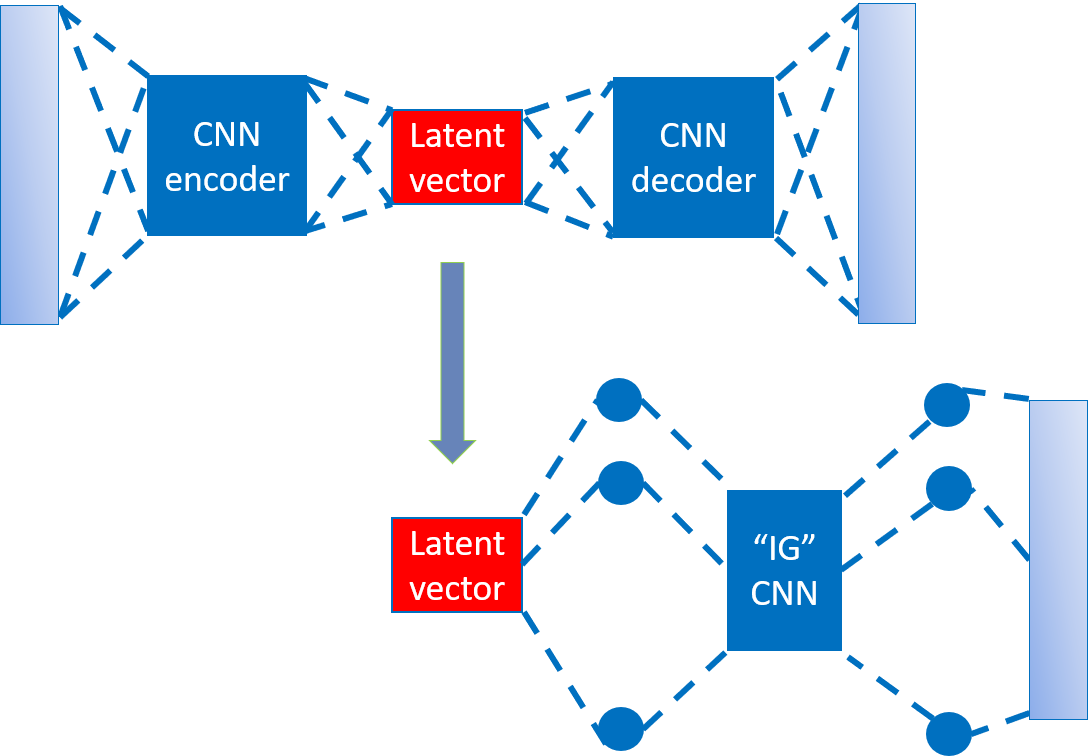}
\caption{The AEIG framework}
\label{fig: AEIG framework}
\end{figure}

By combining the AE network in Section \ref{AE} and IG based network in Section \ref{IG}, the whole framework is constructed, as shown in Fig. \ref{fig: AEIG framework}. The AE will learn and compress the input power maps into compact vectors and the IG will solve the heat equations corresponding to the compact vectors during training. The heat equations for all input power maps are solved in one training process. After the training procedure is completed, besides the power maps seen by the network at training time, variations of these power maps unseen to the network are also used to test the generalization capability of the proposed framework.

\section {Use cases} \label{section_use_cases}
\subsection{Solving the heat equation for sinusoidal power maps}\label{section: sinusoidal_power_map}
\begin{figure}[htbp!]
\centering
\includegraphics[scale=0.4]{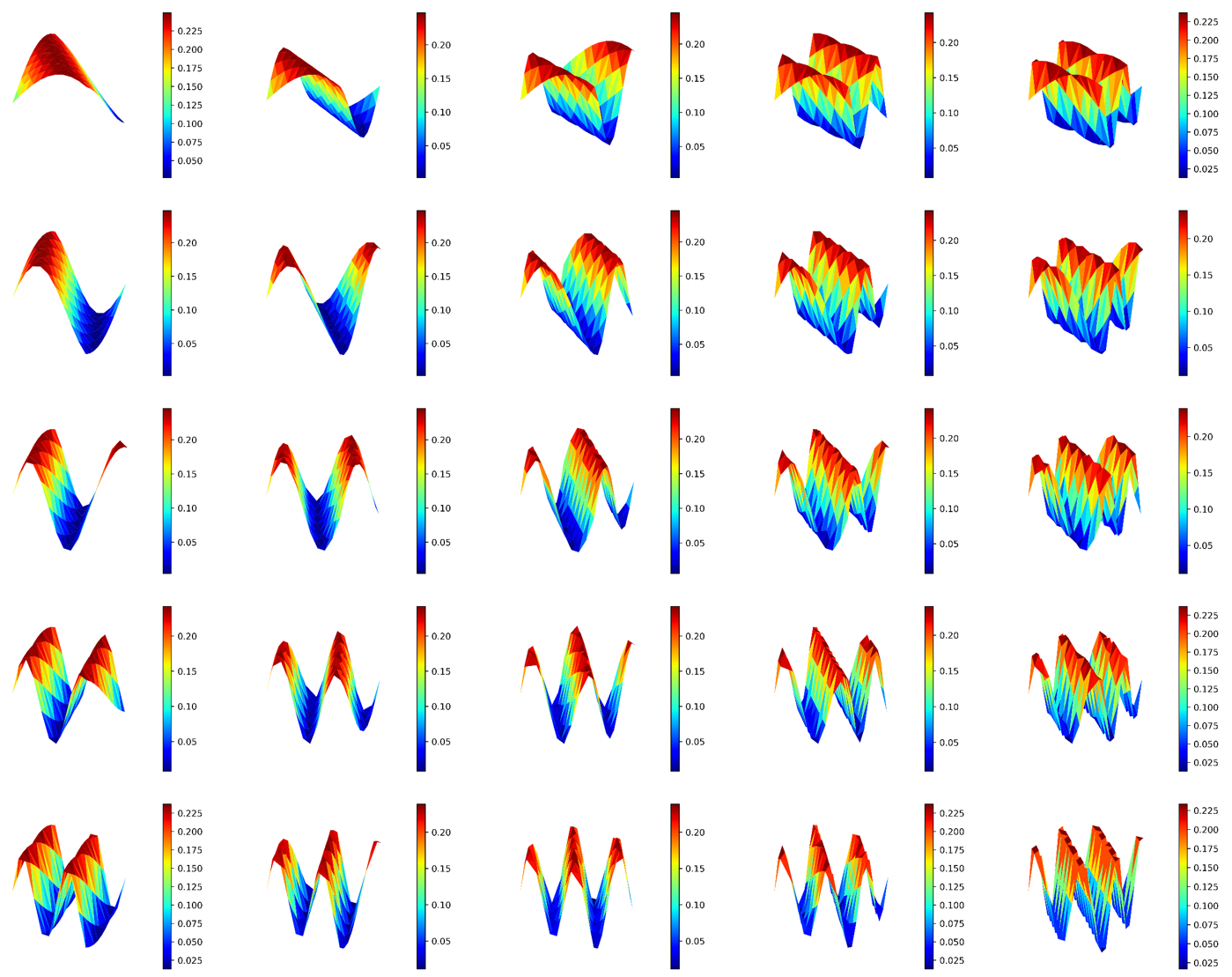}
\caption{Sinusoidal power maps}
\label{fig: Sinusoidal power maps}
\end{figure}

The established framework is first tested on solving sinusoidal power maps, which is described in Eq. \ref{sinusoidal equ}:

\begin{equation}
power(x, y) = C sin(2\pi(ax + by)) + C
\label{sinusoidal equ}
\end{equation}
where $C$ is set to be 0.125,  $a$ and $b$ are both integers in range $[0, 4]$. In total, there are 25 combinations of $a$ and $b$ values and thus creating 25 distinct power maps. The visualization of these power maps is shown in Fig. \ref{fig: Sinusoidal power maps}.

\begin{figure}[htbp!]
\centering
\includegraphics[scale=0.5]{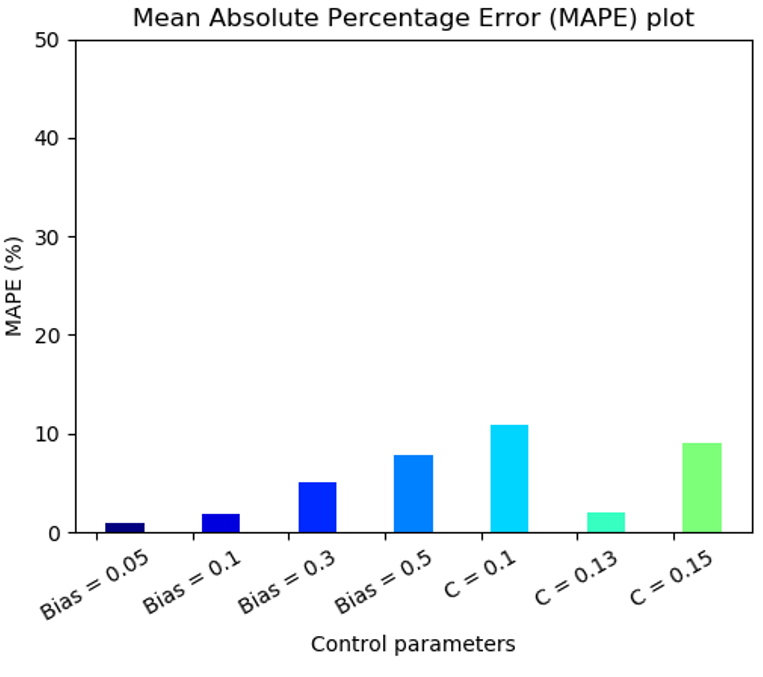}
\caption{Mean absolute percentage error plot for test data sets}
\label{fig: Mean absolute percentage error plot for sinusoidal test data sets}
\end{figure}

After the network training is finished, the network temperature predictions for the 25 training power maps are compared with their corresponding numerical solutions from a numerical solver. Mean absolute percentage error (MAPE) is used as the evaluation metric and the overall MAPE is $0.15\%$: 

Then the trained network is tested on 175 unseen power maps (7 groups in total, 25 power maps in each group). In each of the 7 test sets, only one modification on the control variables is activated. To be specific, either $C$ is modified to be $C'$ taking values in $[0.1, 0.13, 0.15]$ or biases of $[0.05, 0.1, 0.3, 0.5]$ are added to both $a$ and $b$ to form $a'$ and $b'$. The overall analysis on the MAPE error for the test data sets are depicted in Fig. \ref{fig: Mean absolute percentage error plot for sinusoidal test data sets}. For notational convenience, we refer "training points" as the power maps in the training data sets and "test points" as the power maps in the test data sets. The "test points" are close to the "training points" if the absolute differences between parameters $C'$, $a'$, $b'$and $C$, $a$, $b$ are small correspondingly. It can be observed from the plot that the network makes predictions of low MAPE as long as the "test points" are close to the "training points". Even though the error would increase if the "test points" are further away from the "training points", this issue can be alleviated if the "training points" can uniformly cover the complete anticipated field of the unknown inputs, which can be observed from later use cases.

\subsection{Solving the heat equation for exponential power maps}\label{exponential_power_map}
\begin{figure}[htbp!]
\centering
\includegraphics[scale=0.4]{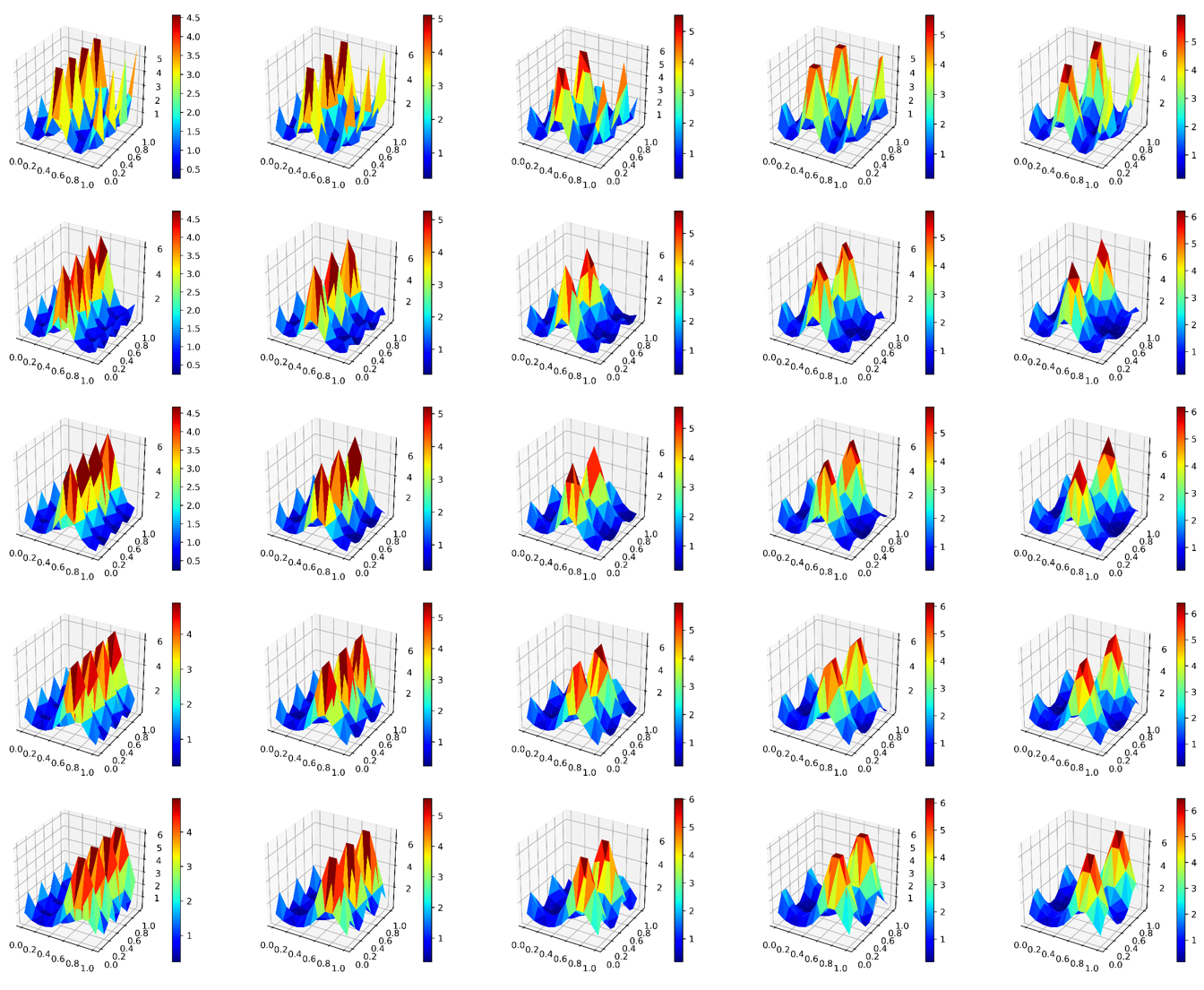}
\caption{Exponential power maps}
\label{fig: Exponential power maps}
\end{figure}

In this test, exponential power maps, which is defined in Eq. \ref{exponential equ}, are created.

\begin{equation}
power(x, y) = e^{-(sin(\frac{2\pi x}{0.5 + 0.1(a + bias)}) + sin(\frac{2\pi y}{0.5 + 0.1(b + bias)}))}
\label{exponential equ}
\end{equation}

where $a$ and $b$ are both integers in range $[0, 4]$ and $bias$ is set to be 0. In total, there are 25 combinations of $a$ and $b$ values and thus 25 different power maps are designed. The visualization of these power maps is shown in Fig. \ref{fig: Exponential power maps}.

After the network has been trained with above mentioned 25 exponential power maps, the temperature predictions it made are validated against numerical solver's results. The MAPE is calculated for each case and the overall MAPE is $0.15\%$.

\begin{figure}[htbp!]
\centering
\includegraphics[scale=0.5]{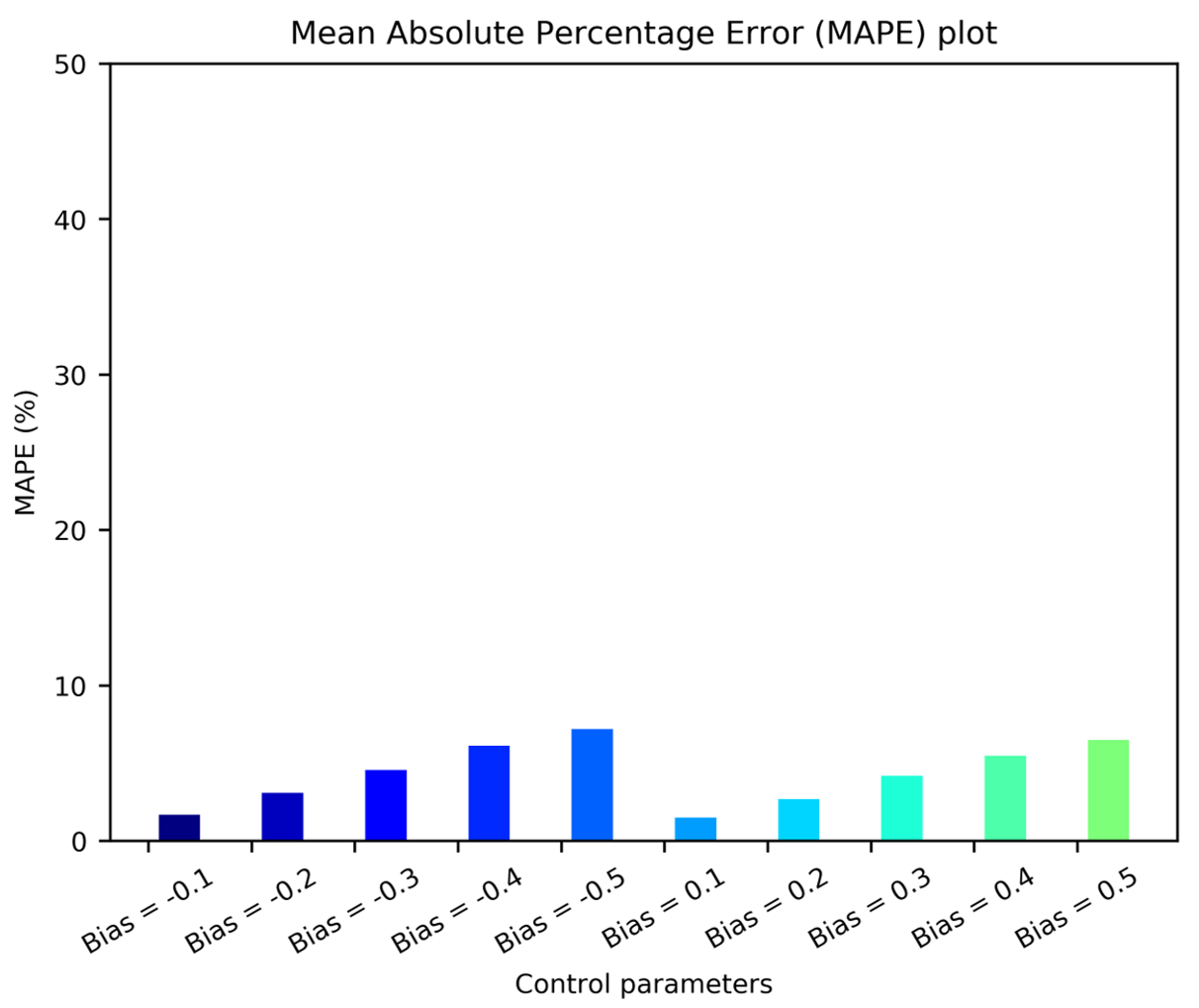}
\caption{Mean absolute percentage error plot for test data sets}
\label{fig: Mean absolute percentage error plot for exp test data sets}
\end{figure}

Similar to Section \ref{section: sinusoidal_power_map}, the trained network is tested on 250 unseen exponential power maps (10 groups in total, 25 power maps in each group) where biases of $[-0.5, -0.4, -0.3, -0.2, -0.1, 0.1, 0.2, 0.3, 0.4, 0.5]$ are added to $a$ and $b$. The MAPE error for the test data sets is investigated and given in Fig. \ref{fig: Mean absolute percentage error plot for exp test data sets}. Similar patterns can be observed that the network makes predictions of low MAPE for "test points" close to the "training points". The prediction accuracy is expected to increase as the "training points" cover the complete anticipated field of the unknown inputs.

\subsection{Solving the heat equation for variations of a realistic chip power map}\label{section: chip_power_map}
\begin{figure}[htbp!]
\centering
\includegraphics[scale=0.3]{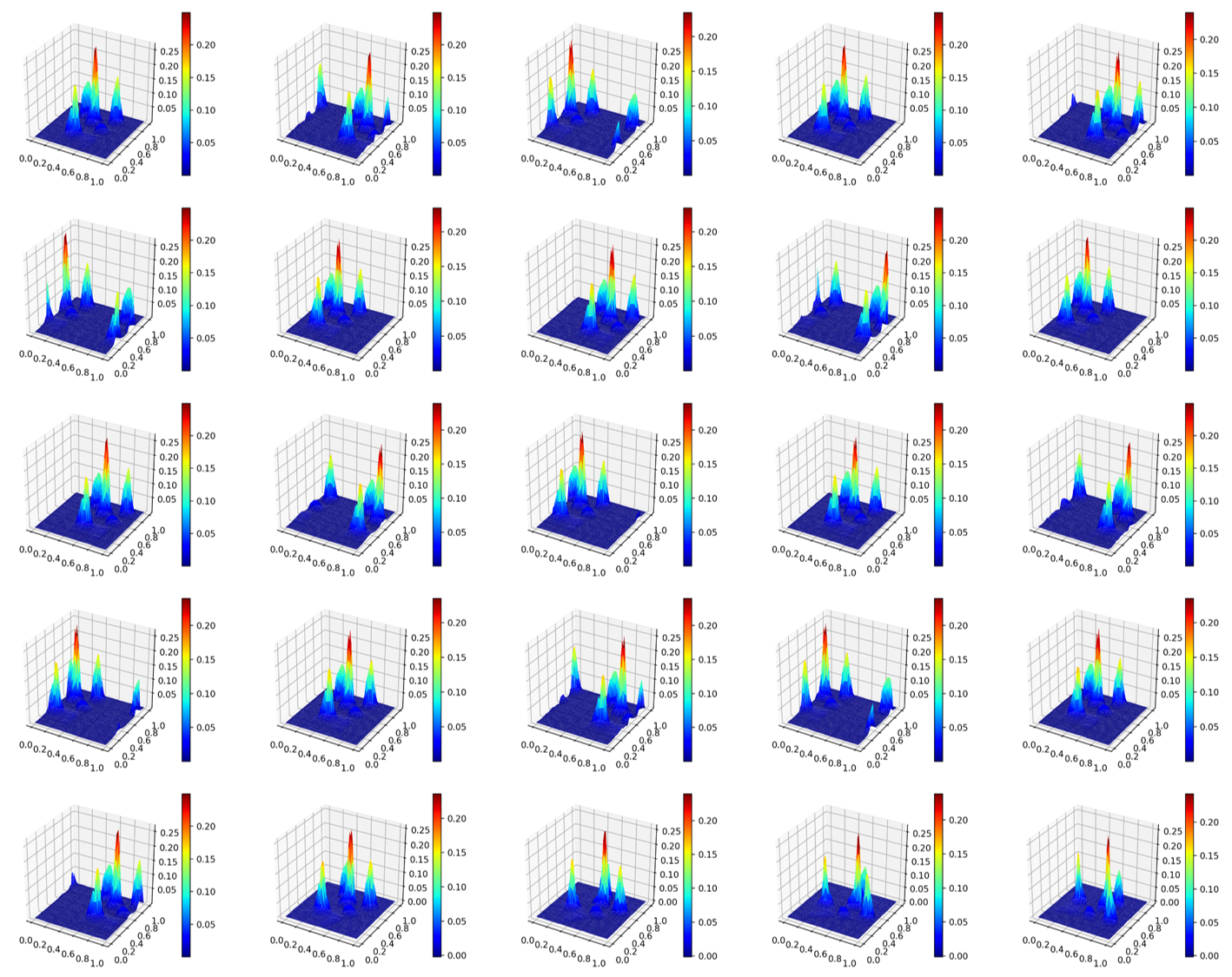}
\caption{Examples for some variations of a realistic chip power map}
\label{fig: scbu power maps}
\end{figure}

As mentioned in Section \ref{section: artificial_power_map}, one realistic chip power map is obtained from a certain chip design. 41 artificial power maps can be created by aforementioned image augmentation techniques, such as translation and rotation of original power map by $c *step$ where $c$ is an integer in range $[1, 20]$ and $step$ is 20. These created power maps are used as the training data for the presented framework. Some examples of these training power maps are provided in Fig. \ref{fig: scbu power maps}.

The equations for the power maps in above data set are solved by the IG based network during its training process. After the training is finished, the network gives the temperature profile predictions. A second-order finite difference numerical solver is utilized to obtain the ground truth results to evaluate the network predictions and the overall MAPE for all the training power maps is $0.19\%$.

\begin{figure}[htbp!]
\centering
\includegraphics[scale=0.5]{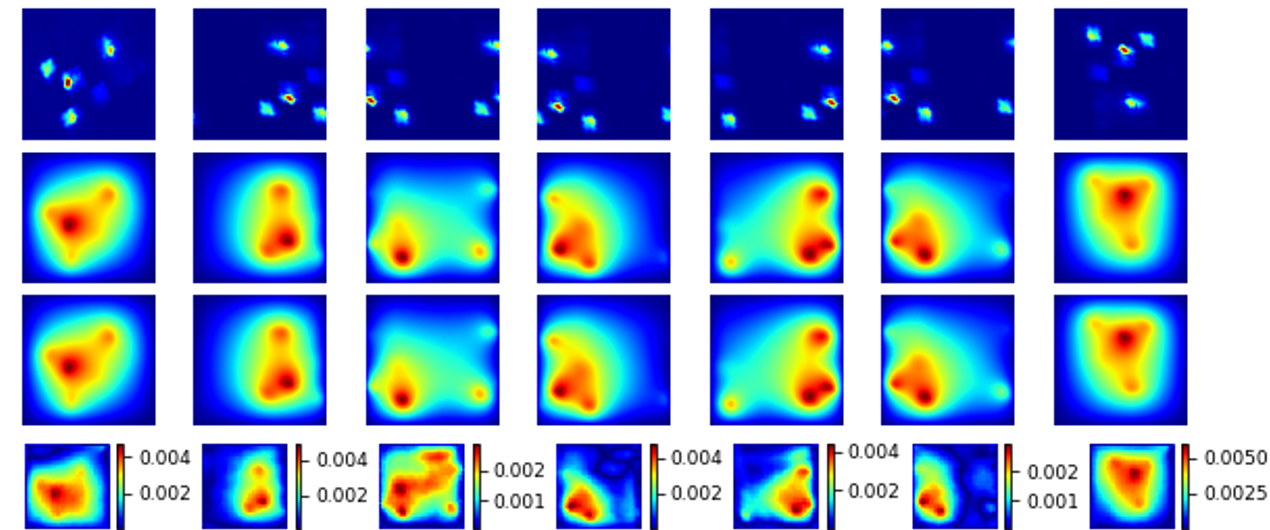}
\caption{An example of network prediction on test data sets}
\label{fig: scbu_pred_example}
\end{figure}

Then the trained network is tested on data sets generated by rotating and shifting the original power maps by  $c' *step'$ where $c'$ is an integer in range $[1, 20]$ and $step'$ can be an integer in range $[1, 20]$. In total, there are 217 distinct unseen power maps. An example of the performance of network predictions is given in Fig. \ref{fig: scbu_pred_example}, with the first row to be the test power map, the second row is the corresponding ground truth solution from numerical solver, the third row is the network prediction and the last row shows the MAPE for each of the network prediction.

To further quantify the effectiveness of the trained network, the network predictions for all the test power maps are compared with solutions from a numerical solver and the overall MAPE for all the test data sets is $0.4\%$.

\subsection{Solving 3D heat equation with convection for variations of a realistic chip power map}\label{section: chip_power_map_3D}
Similar to Section \ref{section: chip_power_map}, 41 artificial power maps are created by applying image augmentation techniques on the given realistic chip power map. The same translation and rotation methods mentioned in Section \ref{section: chip_power_map} are performed, which is rotating or shifting the original power map by $c *step$ where $c$ is an integer in range $[1, 20]$ and $step$ is 20. These created power maps are placed in the center plane of a cube to simulate a simplified chip device with different power maps.

\begin{figure}[htbp!]
\centering
\includegraphics[scale=0.3]{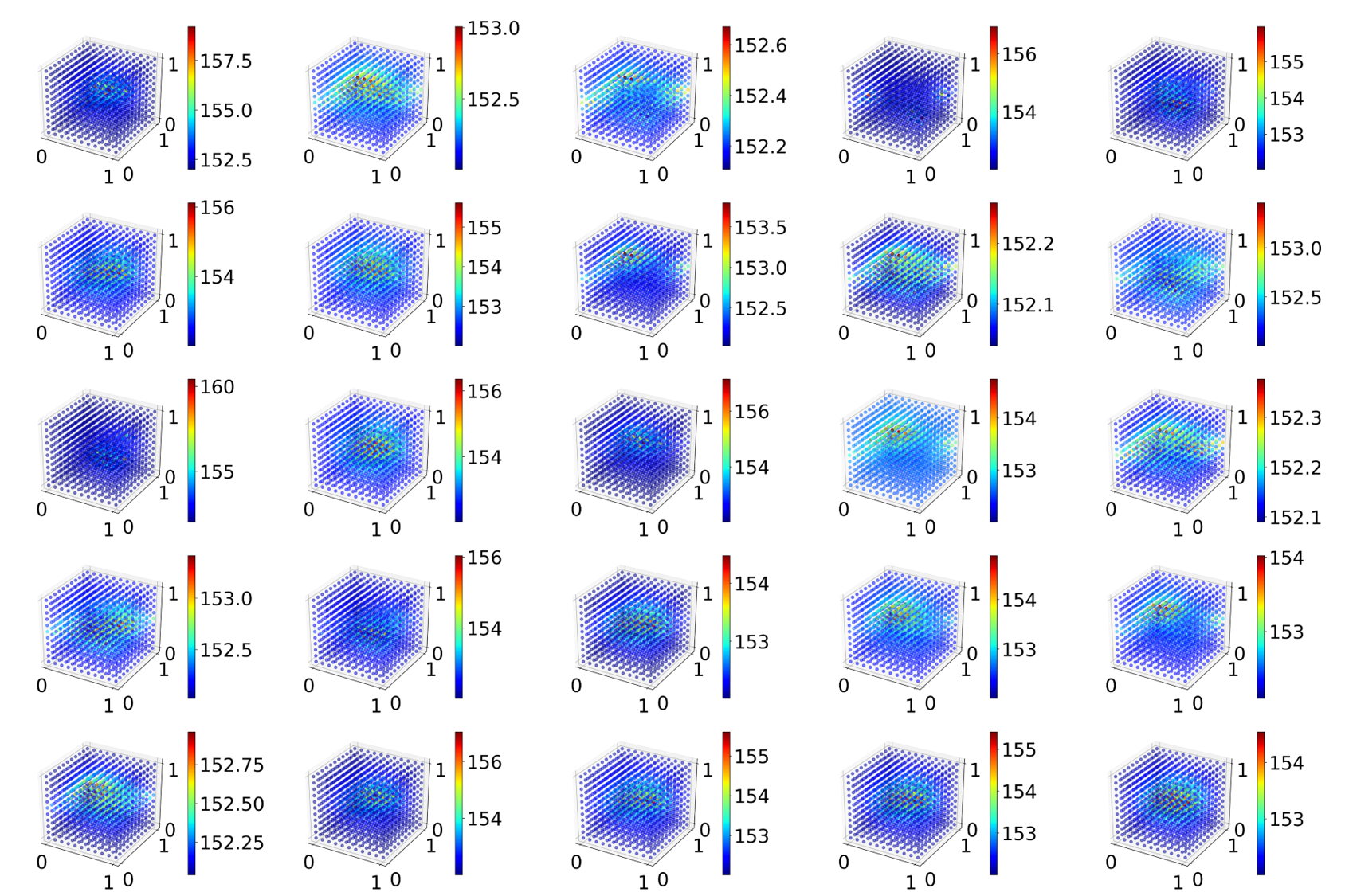}
\caption{Temperature predictions for 3D power maps}
\label{fig: scbu power maps pred_3D}
\end{figure}

The temperature profile for the cubes with different power maps are calculated during the training process and presented in Fig. \ref{fig: scbu power maps pred_3D}. To verify the predictions from the network, the same problems are solved by numerical solvers and the overall MAPE for all the predictions on the training power map is $0.11\%$. 

Finally, the trained network is tested using 188 distinct unseen power maps. All the settings in this section are similar to those  in Section \ref{section: chip_power_map} except that the created 2D power maps are placed in the center plane of 3D cubes and a 3D heat equation with convection, which is defined by Eq. \ref{heat equation with source 3D} is solved. Similarly, the ground truth solutions for these 3D power maps are obtained from a numerical solver and compared with the  network predictions. The overall MAPE for all the test data sets is $0.14\%$. The small MAPE shows the effectiveness of the network's generalization to unseen 3D power maps.

\section {Conclusion} 
This paper presents a framework for solving heat equations on chip in 2D and 3D. The proposed framework can not only solve heat equations with different source terms (distinct power maps) at the same time, it can also generalize the learned physics to heat equations with new source terms. The framework consists of an AE and an IG based network. The AE will learn and compress the source terms of the heat equations and the IG based network will solve heat equations with different source terms during one training process. The effectiveness of the framework on solving sinusoidal, exponential and typical 2D and 3D chip power maps is verified. The potential of using the trained framework on solving heat equations with unseen source terms is tested as well and the low MAPE proves the effectiveness of the proposed framework. It is also worth noting that to obtain reasonable results, the AE network should be fully trained with a set of training data which covers the complete anticipated space of the unknown inputs. The future work would be extending the inference capability of the proposed framework.

\bibliographystyle{unsrtnat}
\bibliography{main}

\begin{thebibliography}{38}
\providecommand{\natexlab}[1]{#1}
\providecommand{\url}[1]{\texttt{#1}}
\expandafter\ifx\csname urlstyle\endcsname\relax
  \providecommand{\doi}[1]{doi: #1}\else
  \providecommand{\doi}{doi: \begingroup \urlstyle{rm}\Url}\fi

\bibitem[He et~al.(2016)He, Lu, Ma, Cao, Shen, and Yu]{he2016joint}
Lifang He, Chun-Ta Lu, Jiaqi Ma, Jianping Cao, Linlin Shen, and Philip~S Yu.
\newblock Joint community and structural hole spanner detection via harmonic
  modularity.
\newblock In \emph{Proceedings of the 22nd ACM SIGKDD International Conference
  on Knowledge Discovery and Data Mining}, pages 875--884, 2016.

\bibitem[Xia et~al.(2018)Xia, Zhang, Yan, Chang, and Yu]{xia2018zero}
Congying Xia, Chenwei Zhang, Xiaohui Yan, Yi~Chang, and Philip~S Yu.
\newblock Zero-shot user intent detection via capsule neural networks.
\newblock \emph{arXiv preprint arXiv:1809.00385}, 2018.

\bibitem[He et~al.(2019)He, Yang, and Pan]{he2019machine}
Haiyang He, Yang Yang, and Yayue Pan.
\newblock Machine learning for continuous liquid interface production: Printing
  speed modelling.
\newblock \emph{Journal of Manufacturing Systems}, 50:\penalty0 236--246, 2019.

\bibitem[Huang et~al.(2020)Huang, Segura, Wang, Zhao, Sun, and
  Zhou]{huang2020unsupervised}
Jida Huang, Luis~Javier Segura, Tianjiao Wang, Guanglei Zhao, Hongyue Sun, and
  Chi Zhou.
\newblock Unsupervised learning for the droplet evolution prediction and
  process dynamics understanding in inkjet printing.
\newblock \emph{Additive Manufacturing}, page 101197, 2020.

\bibitem[Yang et~al.(2019)Yang, He, and Li]{yang2019new}
Yiran Yang, Miao He, and Lin Li.
\newblock A new machine learning based geometry feature extraction approach for
  energy consumption estimation in mask image projection stereolithography.
\newblock \emph{Procedia CIRP}, 80:\penalty0 741--745, 2019.

\bibitem[Psichogios and Ungar(1992)]{psichogios1992hybrid}
Dimitris~C Psichogios and Lyle~H Ungar.
\newblock A hybrid neural network-first principles approach to process
  modeling.
\newblock \emph{AIChE Journal}, 38\penalty0 (10):\penalty0 1499--1511, 1992.

\bibitem[Dissanayake and Phan-Thien(1994)]{dissanayake1994neural}
MWMG Dissanayake and N~Phan-Thien.
\newblock Neural-network-based approximations for solving partial differential
  equations.
\newblock \emph{communications in Numerical Methods in Engineering},
  10\penalty0 (3):\penalty0 195--201, 1994.

\bibitem[Lagaris et~al.(1998)Lagaris, Likas, and
  Fotiadis]{lagaris1998artificial}
Isaac~E Lagaris, Aristidis Likas, and Dimitrios~I Fotiadis.
\newblock Artificial neural networks for solving ordinary and partial
  differential equations.
\newblock \emph{IEEE transactions on neural networks}, 9\penalty0 (5):\penalty0
  987--1000, 1998.

\bibitem[Wu and Xiu(2020)]{wu2020data}
Kailiang Wu and Dongbin Xiu.
\newblock Data-driven deep learning of partial differential equations in modal
  space.
\newblock \emph{Journal of Computational Physics}, 408:\penalty0 109307, 2020.

\bibitem[Shirvany et~al.(2009)Shirvany, Hayati, and
  Moradian]{shirvany2009multilayer}
Yazdan Shirvany, Mohsen Hayati, and Rostam Moradian.
\newblock Multilayer perceptron neural networks with novel unsupervised
  training method for numerical solution of the partial differential equations.
\newblock \emph{Applied Soft Computing}, 9\penalty0 (1):\penalty0 20--29, 2009.

\bibitem[Kaverzneva et~al.(2019)Kaverzneva, Malykhina, and
  Tarkhov]{kaverzneva2019differential}
Tatiana~T Kaverzneva, Galina~F Malykhina, and Dmitriy~A Tarkhov.
\newblock From differential equations to multilayer neural network models.
\newblock In \emph{International Symposium on Neural Networks}, pages 19--27.
  Springer, 2019.

\bibitem[He et~al.(2000)He, Reif, and Unbehauen]{he2000multilayer}
Shouling He, Konrad Reif, and Rolf Unbehauen.
\newblock Multilayer neural networks for solving a class of partial
  differential equations.
\newblock \emph{Neural networks}, 13\penalty0 (3):\penalty0 385--396, 2000.

\bibitem[Sirignano and Spiliopoulos(2018)]{sirignano2018dgm}
Justin Sirignano and Konstantinos Spiliopoulos.
\newblock Dgm: A deep learning algorithm for solving partial differential
  equations.
\newblock \emph{Journal of Computational Physics}, 375:\penalty0 1339--1364,
  2018.

\bibitem[Winovich et~al.(2019)Winovich, Ramani, and Lin]{winovich2019convpde}
Nick Winovich, Karthik Ramani, and Guang Lin.
\newblock Convpde-uq: Convolutional neural networks with quantified uncertainty
  for heterogeneous elliptic partial differential equations on varied domains.
\newblock \emph{Journal of Computational Physics}, 394:\penalty0 263--279,
  2019.

\bibitem[Ranade et~al.(2020)Ranade, Hill, and
  Pathak]{ranade2020discretizationnet}
Rishikesh Ranade, Chris Hill, and Jay Pathak.
\newblock Discretizationnet: A machine-learning based solver for navier-stokes
  equations using finite volume discretization.
\newblock \emph{arXiv preprint arXiv:2005.08357}, 2020.

\bibitem[Baydin et~al.(2017)Baydin, Pearlmutter, Radul, and
  Siskind]{baydin2017automatic}
At{\i}l{\i}m~G{\"u}nes Baydin, Barak~A Pearlmutter, Alexey~Andreyevich Radul,
  and Jeffrey~Mark Siskind.
\newblock Automatic differentiation in machine learning: a survey.
\newblock \emph{The Journal of Machine Learning Research}, 18\penalty0
  (1):\penalty0 5595--5637, 2017.

\bibitem[Raissi et~al.(2017)Raissi, Perdikaris, and
  Karniadakis]{raissi2017physics}
Maziar Raissi, Paris Perdikaris, and George~Em Karniadakis.
\newblock Physics informed deep learning (part i): Data-driven solutions of
  nonlinear partial differential equations.
\newblock \emph{arXiv preprint arXiv:1711.10561}, 2017.

\bibitem[Raissi(2018)]{raissi2018forward}
Maziar Raissi.
\newblock Forward-backward stochastic neural networks: Deep learning of
  high-dimensional partial differential equations.
\newblock \emph{arXiv preprint arXiv:1804.07010}, 2018.

\bibitem[Raissi et~al.(2019)Raissi, Perdikaris, and
  Karniadakis]{raissi2019physics}
Maziar Raissi, Paris Perdikaris, and George~E Karniadakis.
\newblock Physics-informed neural networks: A deep learning framework for
  solving forward and inverse problems involving nonlinear partial differential
  equations.
\newblock \emph{Journal of Computational Physics}, 378:\penalty0 686--707,
  2019.

\bibitem[Lu et~al.(2019{\natexlab{a}})Lu, Meng, Mao, and
  Karniadakis]{lu2019deepxde}
Lu~Lu, Xuhui Meng, Zhiping Mao, and George~E Karniadakis.
\newblock Deepxde: A deep learning library for solving differential equations.
\newblock \emph{arXiv preprint arXiv:1907.04502}, 2019{\natexlab{a}}.

\bibitem[Deng and Hwang(2006)]{deng2006applying}
S~Deng and Y~Hwang.
\newblock Applying neural networks to the solution of forward and inverse heat
  conduction problems.
\newblock \emph{International Journal of Heat and Mass Transfer}, 49\penalty0
  (25-26):\penalty0 4732--4750, 2006.

\bibitem[Mishra(2018)]{mishra2018machine}
Siddhartha Mishra.
\newblock A machine learning framework for data driven acceleration of
  computations of differential equations.
\newblock \emph{arXiv preprint arXiv:1807.09519}, 2018.

\bibitem[Berg and Nystr{\"o}m(2018)]{berg2018unified}
Jens Berg and Kaj Nystr{\"o}m.
\newblock A unified deep artificial neural network approach to partial
  differential equations in complex geometries.
\newblock \emph{Neurocomputing}, 317:\penalty0 28--41, 2018.

\bibitem[Liu et~al.(2019)Liu, Yang, and Cai]{liu2019neural}
Zeyu Liu, Yantao Yang, and Qingdong Cai.
\newblock Neural network as a function approximator and its application in
  solving differential equations.
\newblock \emph{Applied Mathematics and Mechanics}, 40\penalty0 (2):\penalty0
  237--248, 2019.

\bibitem[Wen et~al.(2020{\natexlab{a}})Wen, Pan, Chang, Chuang, Xia, Zhu,
  Kumar, Yang, Srinivasan, and Li]{Jimin}
Jimin Wen, Stephen Pan, Norman Chang, Wen-Tze Chuang, Wenbo Xia, Deqi Zhu,
  Akhilesh Kumar, En-Cih Yang, Karthik Srinivasan, and Ying-Shun Li.
\newblock Dnn-based fast static on-chip thermal solver.
\newblock In \emph{SEMI-THERM 36}, 2020{\natexlab{a}}.

\bibitem[Wen et~al.(2020{\natexlab{b}})Wen, Pan, Chang, Chuang, Xia, Zhu,
  Kumar, Yang, Srinivasan, and Li]{JiminGTC}
Jimin Wen, Stephen Pan, Norman Chang, Wen-Tze Chuang, Wenbo Xia, Deqi Zhu,
  Akhilesh Kumar, En-Cih Yang, Karthik Srinivasan, and Ying-Shun Li.
\newblock A data-driven dnn-based fast chip thermal solver.
\newblock In \emph{GTC}, 2020{\natexlab{b}}.

\bibitem[Cs{\'a}ji et~al.(2001)]{csaji2001approximation}
Bal{\'a}zs~Csan{\'a}d Cs{\'a}ji et~al.
\newblock Approximation with artificial neural networks.
\newblock \emph{Faculty of Sciences, Etvs Lornd University, Hungary},
  24\penalty0 (48):\penalty0 7, 2001.

\bibitem[Hornik et~al.(1989)Hornik, Stinchcombe, White,
  et~al.]{hornik1989multilayer}
Kurt Hornik, Maxwell Stinchcombe, Halbert White, et~al.
\newblock Multilayer feedforward networks are universal approximators.
\newblock \emph{Neural networks}, 2\penalty0 (5):\penalty0 359--366, 1989.

\bibitem[Haykin(1994)]{haykin1994neural}
Simon Haykin.
\newblock \emph{Neural networks: a comprehensive foundation}.
\newblock Prentice Hall PTR, 1994.

\bibitem[Hassoun et~al.(1995)]{hassoun1995fundamentals}
Mohamad~H Hassoun et~al.
\newblock \emph{Fundamentals of artificial neural networks}.
\newblock MIT press, 1995.

\bibitem[Kramer(1991)]{kramer1991nonlinear}
Mark~A Kramer.
\newblock Nonlinear principal component analysis using autoassociative neural
  networks.
\newblock \emph{AIChE journal}, 37\penalty0 (2):\penalty0 233--243, 1991.

\bibitem[Rudy et~al.(2019)Rudy, Alla, Brunton, and Kutz]{rudy2019data}
Samuel Rudy, Alessandro Alla, Steven~L Brunton, and J~Nathan Kutz.
\newblock Data-driven identification of parametric partial differential
  equations.
\newblock \emph{SIAM Journal on Applied Dynamical Systems}, 18\penalty0
  (2):\penalty0 643--660, 2019.

\bibitem[Lu et~al.(2019{\natexlab{b}})Lu, Kim, and
  Solja{\v{c}}i{\'c}]{lu2019extracting}
Peter~Y Lu, Samuel Kim, and Marin Solja{\v{c}}i{\'c}.
\newblock Extracting interpretable physical parameters from spatiotemporal
  systems using unsupervised learning.
\newblock \emph{arXiv preprint arXiv:1907.06011}, 2019{\natexlab{b}}.

\bibitem[Long et~al.(2017)Long, Lu, Ma, and Dong]{long2017pde}
Zichao Long, Yiping Lu, Xianzhong Ma, and Bin Dong.
\newblock Pde-net: Learning pdes from data.
\newblock \emph{arXiv preprint arXiv:1710.09668}, 2017.

\bibitem[Erichson et~al.(2019)Erichson, Muehlebach, and
  Mahoney]{erichson2019physics}
N~Benjamin Erichson, Michael Muehlebach, and Michael~W Mahoney.
\newblock Physics-informed autoencoders for lyapunov-stable fluid flow
  prediction.
\newblock \emph{arXiv preprint arXiv:1905.10866}, 2019.

\bibitem[Goh et~al.(2019)Goh, Sheriffdeen, and Bui-Thanh]{goh2019solving}
Hwan Goh, Sheroze Sheriffdeen, and Tan Bui-Thanh.
\newblock Solving forward and inverse problems using autoencoders.
\newblock \emph{arXiv preprint arXiv:1912.04212}, 2019.

\bibitem[Pakravan et~al.(2020)Pakravan, Mistani, Aragon-Calvo, and
  Gibou]{pakravan2020solving}
Samira Pakravan, Pouria~A Mistani, Miguel~Angel Aragon-Calvo, and Frederic
  Gibou.
\newblock Solving inverse-pde problems with physics-aware neural networks.
\newblock \emph{arXiv preprint arXiv:2001.03608}, 2020.

\bibitem[Champion et~al.(2019)Champion, Lusch, Kutz, and
  Brunton]{champion2019data}
Kathleen Champion, Bethany Lusch, J~Nathan Kutz, and Steven~L Brunton.
\newblock Data-driven discovery of coordinates and governing equations.
\newblock \emph{Proceedings of the National Academy of Sciences}, 116\penalty0
  (45):\penalty0 22445--22451, 2019.

\end{thebibliography}
\end{document}